%% file: main.tex
\documentclass{article}

\usepackage{times}
\usepackage{graphicx} 
\usepackage{subfigure} 
\usepackage{amsmath}
\usepackage{amsfonts}
\usepackage{amssymb,amsthm}
\usepackage{bbm}
\usepackage{enumitem}
\usepackage{tabularx}
\usepackage{multirow}
\usepackage{caption}
\usepackage{float}
\floatstyle{plaintop}
\restylefloat{table}
\usepackage{natbib}

\usepackage{geometry}
\newgeometry{top=4.4cm}

\usepackage{algorithm}
\usepackage{algorithmic}
\usepackage{math_macro}
\usepackage{hyperref}

\usepackage[accepted]{icml2017} 

\icmltitlerunning{Deriving Neural Architectures from Sequence and Graph Kernels}

\begin{document} 

\twocolumn[
\icmltitle{Deriving Neural Architectures from Sequence and Graph Kernels}



\icmlsetsymbol{equal}{*}

\begin{icmlauthorlist}
\icmlauthor{Tao Lei*}{mit}
\icmlauthor{Wengong Jin*}{mit}
\icmlauthor{Regina Barzilay}{mit}
\icmlauthor{Tommi Jaakkola}{mit}

\end{icmlauthorlist}

\icmlaffiliation{mit}{MIT Computer Science \& Artificial Intelligence Laboratory}
\icmlcorrespondingauthor{Tao Lei}{taolei@csail.mit.edu}
\icmlcorrespondingauthor{Wengong Jin}{wengong@csail.mit.edu}

\icmlkeywords{kernel, deep learning}

\vskip 0.3in
]
\printAffiliationsAndNotice{\icmlEqualContribution} 
\allowdisplaybreaks
\input{abstract}
\input{intro}
\input{seq_rcnn}

\input{graph_rcnn}
\input{gated_rcnn}
\input{rel_work}

\input{experiment}
\input{conclusion}


\bibliography{main}
\bibliographystyle{icml2017}

\newpage
\input{appendix}

\end{document}

%% file: abstract.tex
\begin{abstract}
The design of neural architectures for structured objects is typically guided by experimental insights rather than a formal process.
In this work, we appeal to kernels over combinatorial structures, such as sequences and graphs, to derive appropriate neural operations.
We introduce a class of deep recurrent neural operations and formally characterize their associated kernel spaces. Our recurrent modules compare the input to virtual reference objects (cf. filters in CNN) via the kernels. 
Similar to traditional neural operations, these reference objects are parameterized and directly optimized in end-to-end training.
We empirically evaluate the proposed class of neural architectures on standard applications such as language modeling and molecular graph regression, achieving state-of-the-art results across these applications.
\end{abstract}

%% file: intro.tex
\section{Introduction}
Many recent studies focus on designing novel neural architectures for structured data such as sequences or annotated graphs. For instance, LSTM~\cite{hochreiter1997long}, GRU~\cite{chung2014empirical} and other complex recurrent units~\cite{zoph2016neural} can be easily adapted to embed structured objects such as sentences~\cite{tai2015improved} or 
molecules~\cite{li2015gated,dai2016discriminative} into vector spaces suitable for later processing by standard predictive methods. The embedding algorithms are typically integrated into an end-to-end trainable architecture so as to tailor the learnable embeddings directly to the task at hand. 

The embedding process itself is characterized by a sequence operations summarized in a structure known as the computational graph. Each node in the computational graph identifies the unit/mapping applied while the arcs specify the relative arrangement/order of operations. The process of designing such computational graphs or associated operations for classes of objects is often guided by insights and expertise rather than a formal process.

Recent work has substantially narrowed the gap between desirable computational operations associated with objects and how their representations are acquired. For example, value iteration calculations can be folded into convolutional architectures so as to optimize the representations to facilitate planning~\cite{tamar2016value}. Similarly, inference calculations in graphical models about latent states of variables such as atom characteristics can be directly associated with embedding operations~\cite{dai2016discriminative}. 

We appeal to kernels over combinatorial structures to define the appropriate computational operations. Kernels give rise to well-defined function spaces and possess rules of composition that guide how they can be built from simpler ones. The comparison of objects inherent in kernels is often broken down to elementary relations such as counting of common sub-structures as in
\begin{equation}
\kernel(\chi,\chi') = \sum_{s\in\mathcal{S}} \mathbf{1}[s\in\chi]\mathbf{1}[s\in\chi']
\end{equation}
where $\mathcal{S}$ is the set of possible substructures. For example, in a string kernel~\cite{lodhi2002text}, $\mathcal{S}$ may refer to all possible subsequences while a graph kernel~\cite{vishwanathan2010graph} would deal with possible paths in the graph.  Several studies have highlighted the relation between feed-forward neural architectures and kernels~\cite{hazan2015steps,zhang16l1} but we are unaware of any prior work pertaining to kernels associated with neural architectures for structured objects. 

In this paper, we introduce a class of deep recurrent neural embedding operations and formally characterize their associated kernel spaces. The resulting kernels are parameterized in the sense that the neural operations relate objects of interest to virtual reference objects through kernels. These reference objects are parameterized and readily optimized for end-to-end performance. 

To summarize, the proposed neural architectures, or \emph{Kernel Neural Networks}
\footnote{Code available at \href{https://github.com/taolei87/icml17_knn}{https://github.com/taolei87/icml17\_knn}}
, enjoy the following advantages:
\begin{itemize}[itemsep=0pt]
\vspace{-5pt}
\item The architecture design is grounded in kernel computations.
\item Our neural models remain end-to-end trainable to the task at hand.
\item Resulting architectures demonstrate state-of-the-art performance against strong baselines.
\vspace{-5pt}
\end{itemize}
In the following sections, we will introduce these neural components derived from string and graph kernels, as well as their deep versions.
Due to space limitations, we defer proofs to supplementary material.

%% file: seq_rcnn.tex
\section{From String Kernels to Sequence NNs}
\label{sec:string}
{\bf Notations } We define a sequence (or a string) of tokens (e.g. a sentence) as $\x_{1:L}=\{\x_i\}_{i=1}^{L}$ where $\x_i\in\real^d$ represents its $i^\text{th}$ element and $\vert\x\vert=L$ denotes the length. 
Whenever it is clear from the context, we will omit the subscript and directly use $\x$ (and $\y$) to denote a sequence. For a pair of vectors (or matrices) $\mbf{u}, \mbf{v}$, we denote $\inner{\mbf{u}}{\mbf{v}}=\sum_k u_k v_k$ as their inner product.
For a kernel function $\kernel_i(\cdot,\cdot)$ with subscript $i$, we use $\phi_i(\cdot)$ to denote its underlying mapping, i.e. $\kernel_i(\x,\y)=\inner{\phi_i(\x)}{\phi_i(\y)}=\phi_i(\x)^\top\phi_i(\y)$.

{\bf String Kernel } String kernel measures the similarity between two sequences by counting shared subsequences (see~\citet{lodhi2002text}). For example, let $\x$ and $\y$ be two strings, a bi-gram string kernel $\kernel_2(\x,\y)$ counts the number of bi-grams $(\x_i,\x_j)$ and $(\y_k,\y_l)$ such that $(\x_i,\x_j)=(\y_k,\y_l)$\footnote{We define n-gram as a \textbf{subsequence} of original string (not necessarily consecutive).},
\begin{equation}
\kernel_{2}(\x,\y) = \sum_{\substack{1\leq i<j\leq\vert\x\vert \\ 1\leq k<l\leq\vert\y\vert}}\lambda_{\x,i,j}\,\lambda_{\y,k,l} \,\delta(\x_i,\y_k)\cdot\delta(\x_j,\y_l)
\label{eq:seqkernel}
\end{equation}
where $\lambda_{\x,i,j}, \lambda_{\y,k,l}\in [0,1)$ are context-dependent
weights and $\delta(x,y)$ is an indicator that returns 1 only when $x=y$. 
The weight factors can be realized in various ways. 
For instance, in temporal predictions such as language modeling, substrings
(i.e. patterns) which appear later may have higher impact for prediction. 
Thus a realization $\lambda_{\x,i,j}=\lambda^{\vert\x\vert-i-1}$ and
$\lambda_{\y,k,l}=\lambda^{\vert\y\vert-k-1}$ (penalizing substrings far from the
end) can be used to determine weights given a constant decay factor $\lambda\in (0,1)$.

In our case, each token in the sequence is a vector (such as one-hot encoding of
a word or a feature vector). We shall replace the exact match $\delta(\mbf{u},\mbf{v})$ by the inner product $\inner{\mbf{u}}{\mbf{v}}$. To this end, the kernel function~(\ref{eq:seqkernel}) can be rewritten as,
\begin{align}
&\ \sum_{1\leq i<j\leq\vert\x\vert}\ \sum_{1\leq k<l\leq\vert\y\vert}
\lambda_{\x,i,j}\,\lambda_{\y,k,l} \inner{\x_i}{\y_k}\cdot \inner{\x_j}{\y_l} & \nonumber \\
&=\ \sum_{1\leq i<j\leq\vert\x\vert}\ \sum_{1\leq k<l\leq\vert\y\vert}
\lambda_{\x,i,j}\,\lambda_{\y,k,l} \inner{\x_i\otimes\x_j}{\y_k\otimes\y_l} & \nonumber \\
\ &= \inner{\ \sum_{i<j}\ \lambda^{\vert\x\vert-i-1}\ \x_i\otimes\x_j}{\ \ \sum_{k<l}\
\lambda^{\vert\y\vert-k-1}\ \y_k\otimes\y_l}
\label{eq:seqkernel-mul}
\end{align}
where $\x_i\otimes\x_j\in\real^{d\times d}$ (and similarly $\y_k\otimes\y_l$) is the outer-product. 
In other words, the underlying mapping of kernel $\kernel_2()$ defined above is $\phi_2(\x)=\sum_{1\leq i<j\leq\vert\x\vert} \lambda^{\vert\x\vert-i-1} \x_i\otimes\x_j$. 
Note we could alternatively use a partial additive scoring
$\inner{\x_i}{\y_k}+\inner{\x_j}{\y_l}$, and the kernel function can be
generalized to n-grams when $n\neq 2$. 
Again, we commit to one realization in this section.

{\bf String Kernel NNs }
We introduce a class of recurrent modules whose internal feature states embed the
computation of string kernels. 
The modules \emph{project} kernel mapping $\phi(\x)$ into multi-dimensional vector space (i.e. internal states of recurrent nets).
Owing to the combinatorial structure of $\phi(\x)$, such projection can be realized and factorized via efficient computation.
For the example kernel discussed above, the corresponding neural component is realized as,
\begin{align}
 \mbf{c}_1[t] &= \lambda\cdot \mbf{c}_1[t-1] + \left(\W^{(1)}
 \x_t\right)\nonumber\\
 \mbf{c}_j[t] &= \lambda\cdot \mbf{c}_j[t-1] +
 \left(\mbf{c}_{j-1}[t-1]\odot\W^{(j)}\x_t\right) \nonumber\\
 \mbf{h}[t] &= \sigma(\mbf{c}_n[t]), \qquad\qquad\qquad\qquad\ \  1<j\leq n
 \label{eq:seqnn}
\end{align}
where $\cc_j[t]$ are the pre-activation cell states at word $\x_t$, and $\h[t]$ is the (post-activation) hidden vector. $\cc_j[0]$ is initialized with a zero vector. $\W^{(1)},..,\W^{(n)}$ are weight matrices to be learned from training examples.

The network operates like other RNNs by processing each input token and updating
the internal states.
The elementwise multiplication $\odot$ can be replaced by
addition $+$ (corresponding to the partial additive scoring above).
As a special case, the additive variant becomes a word-level convolutional neural net~\cite{Kim14} when
$\lambda=0$.\footnote{$\h[t]=\sigma(\W^{(1)}\x_{t-n+1}+\cdots +\W^{(n)}\x_t)$ when $\lambda = 0$.}

\begin{figure}[!t!]
\vspace{0.05in}
\includegraphics[width=3in]{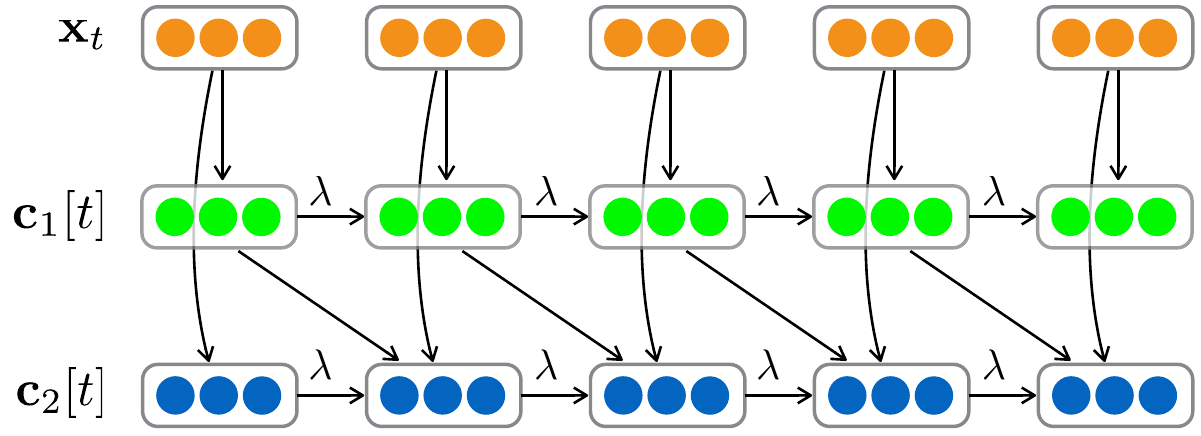}
\caption{An unrolled view of the derived recurrent module for $\kernel_2()$. Horizontal lines denote decayed propagation from $\cc[t-1]$ to $\cc[t]$, while vertical lines represent a linear mapping $\W\x_t$ that is propagated to the internal states $\cc[t]$.}
\end{figure}

\subsection{Single Layer as Kernel Computation}
Now we state how the proposed class embeds string kernel computation. 
For $j\in\{1,..,n\}$, let $c_j[t][i]$ be the i-th entry of state vector $\mbf{c}_j[t]$, $\mbf{w}_i^{(j)}$ represents the i-th row of matrix $\W^{(j)}$. 
Define $\mbf{w}_{i,j}=\{\mbf{w}^{(1)}_i, \mbf{w}^{(2)}_i, ... ,
\mbf{w}^{(j)}_i\}$ as a ``reference sequence'' constructed by taking the i-th row from each matrix $\W^{(1)},..,\W^{(j)}$.
\begin{theo}
Let $\x_{1:t}$ be the prefix of $\x$ consisting of first $t$ tokens, and
$\kernel_{j}$ be the string kernel of $j$-gram shown in
Eq.(\ref{eq:seqkernel-mul}). Then $\cc_j[t][i]$ evaluates kernel function,
\begin{align*}
\cc_j[t][i]\ &=\ \kernel_{j}\left(\x_{1:t}, \mbf{w}_{i,j}\right) = \inner{\phi_j(\x_{1:t})}{\phi_j(\mbf{w}_{i,j})}
\end{align*}
for any $j\in\{1,..,n\}$, $t\in\{1,..,\vert\x\vert\}$. 
\label{theorem:onelayer}
\end{theo}
In other words, the network \emph{embeds sequence similarity computation} by assessing the similarity between the input sequence $\x_{1:t}$ and the reference sequence $\mbf{w}_{i,j}$.
This interpretation is similar to that of CNNs, where each filter is a ``reference
pattern'' to search in the input.
String kernel NN further takes non-consecutive n-gram patterns into consideration (seen from the summation over all n-grams in Eq.(\ref{eq:seqkernel-mul})).

{\bf Applying Non-linear Activation } In practice, a non-linear activation
function such as polynomial or sigmoid-like activation is added to the internal
states to produce the final output state $\h[t]$. It turns out that many
activations are also functions in the reproducing kernel Hilbert space (RKHS) of certain kernel functions
(see~\citet{shalev2011learning,zhang16l1}). 
When this is true, the underlying kernel of $\h[t]$ is the composition of string
kernel and the kernel containing the activation. We give the formal statements below.

\begin{lemma} 
Let $\x$ and $\mbf{w}$ be multi-dimensional vectors with finite norm. Consider the function $f(\x):=\sigma(\mbf{w}^\top \x)$ with non-linear activation $\sigma(\cdot)$. For functions such as polynomials and sigmoid function, there exists kernel functions $\kernel_\sigma(\cdot,\cdot)$ and the underlying mapping $\phi_\sigma(\cdot)$ such that $f(x)$ is in the reproducing kernel Hilbert space of $\kernel_\sigma(\cdot,\cdot)$, i.e.,
\begin{align*}
f(\x)=\sigma(\mbf{w}^\top\x) = \inner{\phi_\sigma(\x)}{\psi(\mbf{w})}
\end{align*}
for some mapping $\psi(\mbf{w})$ constructed from $\mbf{w}$.
In particular, $\kernel_\sigma(\x,\y)$ can be the inverse-polynomial kernel $\frac{1}{2-\inner{\x}{\y}}$ for the above activations.
\label{lemma:zhang}
\end{lemma}

\begin{prop}
For one layer string kernel NN with non-linear activation $\sigma(\cdot)$ discussed in Lemma~\ref{lemma:zhang}, 
$\h[t][i]$ as a function of input $\x$ belongs to the RKHS introduced by the composition of $\kernel_\sigma(\cdot,\cdot)$ and string kernel $\kernel_n(\cdot,\cdot)$.
Here a kernel composition $\kernel_{\sigma,n}(\x,\y)$ is defined with the underlying mapping $\x \mapsto \phi_\sigma(\phi_n(\x))$, and hence $\kernel_{\sigma,n}(\x,\y)=\phi_\sigma(\phi_n(\x))\top\phi_\sigma(\phi_n(\y))$.
\label{theorem:onelayer_act}
\end{prop}

Proposition~\ref{theorem:onelayer_act} is the corollary of Lemma~\ref{lemma:zhang} and Theorem~\ref{theorem:onelayer}, since $\h[t][i]=\sigma(\mbf{c}_n[t][i])=\sigma(\kernel_n(\x_{1:t},\mbf{w}_{i,j}))=\inner{\phi_\sigma(\phi_n(\x_{1:t}))}{\tilde{\mbf{w}}_{i,j}}$ and $\phi_\sigma(\phi_n(\cdot))$ is the mapping for the composed kernel. The same proof applies when $\h[t]$ is a linear combination of all $\mbf{c}_i[t]$ since kernel functions are closed under addition.

\subsection{Deep Networks as Deep Kernel Construction} 
\label{sec:deep_seqrcnn}
We now address the case when multiple layers of the same module are stacked to construct deeper networks. That is, the output states $\h^{(l)}[t]$ of the $l$-th layer are fed to the $(l+1)$-th layer as the input sequence. We show that layer stacking corresponds to recursive kernel construction (i.e. $(l+1)$-th kernel is defined on top of $l$-th kernel), which has been proven for feed-forward networks~\cite{zhang16l1}.

We first generalize the sequence kernel definition to enable recursive
construction. Notice that the definition in Eq.(\ref{eq:seqkernel-mul}) uses the
linear kernel (inner product) $\inner{\x_i}{\y_k}$ as a ``subroutine'' to measure the similarity between substructures (e.g. tokens) within the sequences. 
We can therefore replace it with other similarity measures introduced by other
``base kernels''. In particular,
let $\kernel^{(1)}(\cdot,\cdot)$ be the string kernel (associated with a single
layer). The generalized sequence kernel $\kernel^{(l+1)}(\x,\y)$ can be recursively defined as,
\begin{align}
&\sum_{\substack{i<j\\k<l}}\ \lambda_{\x,i,j}\ \lambda_{\y,k,l}
\ \kernel^{(l)}_\sigma(\x_{1:i},\y_{1:k}) \ \kernel^{(l)}_\sigma(\x_{1:j},\y_{1:l})
\ \ = \nonumber\\
&\inner{\sum_{i<j}\phi^{(l)}_\sigma(\x_{1:i})\otimes\phi^{(l)}_\sigma(\x_{1:j})}{\sum_{k<l}\phi^{(l)}_\sigma(\y_{1:k})\otimes\phi^{(l)}_\sigma(\y_{1:l})} \nonumber
\label{eq:sk_generalized}
\end{align}
where $\phi^{(l)}(\cdot)$ denotes the pre-activation mapping of the $l$-th kernel, $\phi^{(l)}_\sigma(\cdot)=\phi_\sigma(\phi^{(l)}(\cdot))$ denotes the underlying (post-activation) mapping for non-linear activation $\sigma(\cdot)$, and $\kernel_\sigma^{(l)}(\cdot,\cdot)$ is the $l$-th post-activation kernel.
Based on this definition, a deeper model can also be interpreted as a kernel computation.

\begin{theo}
Consider a deep string kernel NN with $L$ layers and activation function $\sigma(\cdot)$. Let the final output state $\h^{(L)}[t]=\sigma(\mbf{c}_n^{(L)}[t])$ (or any linear combination of $\{\mbf{c}^{(l)}_i[t]\}, i=1,..,n$). For $l=1,\cdots,L$,
\begin{enumerate}[itemsep=0pt]
\vspace{-5pt}
\item[(i)] $\mbf{c}^{(l)}_n[t][i]$ as a function of input $\x$ belongs to the RKHS of kernel $\kernel^{(l)}(\cdot,\cdot)$;
\item[(ii)] $\h^{(l)}[t][i]$ belongs to the RKHS of kernel $\kernel^{(l)}_\sigma(\cdot,\cdot)$.
\end{enumerate}
\label{theorem:deep_act}
\end{theo}

%% file: graph_rcnn.tex
\section{From Graph Kernels to Graph NNs}
\label{sec:graph}

In the previous section, we encode sequence kernel computation into neural modules and demonstrate possible extensions using different base kernels.
The same ideas apply to other types of kernels and data.
Specifically, we derive neural components for graphs in this section.

{\bf Notations } A graph is defined as $G=(V,E)$, with each vertex $v \in V$ associated with feature vector $\f_v$. The neighbor of node $v$ is denoted as $N(v)$. Following previous notations, for any kernel function $\kernel_*(\cdot,\cdot)$ with underlying mapping $\phi_*(\cdot)$, we use $\kernel_{*,\sigma}(\cdot,\cdot)$ to denote the post-activation kernel induced from the composed underlying mapping $\phi_{*,\sigma}=\phi_\sigma(\phi_*(\cdot))$.

\subsection{Random Walk Kernel NNs}
We start from random walk graph kernels \cite{gartner2003graph}, which count common walks in two graphs. Formally, let $P_n(G)$ be the set of walks $\x=x_1\cdots x_n$, where $\forall i:(x_i,x_{i+1})\in E$.\footnote{A single node could appear multiple times in a walk.} Given two graphs $G$ and $G'$, an $n$-th order random walk graph kernel is defined as:
\begin{equation}
\kernel_W^n(G,G') = \lambda^{n-1} \sum_{\x \in P_n(G)} \sum_{\y \in P_n(G')} \prod_{i=1}^n \inner{\f_{\x_i}}{\f_{\y_i}}
\label{eq:randomwalk}
\end{equation}
where $\f_{x_i}\in \mathbb{R}^d$ is the feature vector of node $\x_i$ in the walk.

Now we show how to realize the above graph kernel with a neural module. Given a graph $G$, the proposed neural module is:
\begin{eqnarray}
\cc_1[v] &=& \W^{(1)} \f_v \nonumber \\
\cc_j[v] &=& \lambda \sum_{u\in N(v)} c_{j-1}[u] \odot \W^{(j)} \f_v \label{eq:rw_nn} \\
\h_G &=& \sigma(\sum_v\cc_n[v]) \qquad\qquad\quad 1<j\leq n\nonumber
\end{eqnarray}
where again $\cc_*[v]$ is the cell state vector of node $v$, and $\h_G$ is the representation of graph $G$ aggregated from node vectors. $\h_G$ could then be used for classification or regression.

Now we show the proposed model embeds the random walk kernel. To show this, construct $L_{n,k}$ as a ``reference walk'' consisting of the row vectors $\{\w^{(1)}_k,\cdots,\w^{(n)}_k\}$ from the parameter matrices.
Here $L_{n,k}=(L_V,L_E)$, where $L_V=\{v_0,v_1,\cdots,v_n\}$, $L_E=\{(v_i,v_{i+1})\}$ and $v_i$'s feature vector is $\w^{(i)}_k$.
We have the following theorem:
\begin{theo}
For any $n\geq 1$, the state value $c_n[v][k]$ (the $k$-th coordinate of $c_n[v]$) satisfies:
$$
\sum_v \cc_n[v][k] = \kernel_W^n(G,L_{n,k})
$$
thus $\sum_v \cc_n[v]$ lies in the RKHS of kernel $\kernel^n_W$. As a corollary, $\h_G$ lies in the RKHS of kernel $K_{W,\sigma}^n()$.
\label{theorem:graph-onelayer}
\end{theo}

\subsection{Unified View of Graph Kernels}
The derivation of the above neural module could be extended to other classes of graph kernels, such as subtree kernels (cf.~\cite{ramon2003expressivity,vishwanathan2010graph}).
Generally speaking, most of these kernel functions factorize graphs into local sub-structures, i.e.
\begin{equation}
\kernel(G,G') = \sum_v\sum_{v'} \kernel_{loc}(v,v')
\end{equation}
where $\kernel_{loc}(v,v')$ measures the similarity between local sub-structures centered at node $v$ and $v'$.

For example, the random walk kernel $K_W^n$ can be equivalently defined with $\kernel_{loc}^n(v,v')=$
$$
\begin{cases}
\displaystyle
\ \inner{\f_v}{\f_{v'}} & \text{if } n=1 \\[0.5ex]
\ \inner{\f_v}{\f_{v'}} \cdot \lambda \sum\limits_{u\in N(v)}\sum\limits_{u'\in N(v')} \kernel_{loc}^{n-1}(u,u') & \text{if } n>1
\end{cases}
$$
Other kernels like subtree kernels could be recursively defined similarly. Therefore, we adopt this unified view of graph kernels for the rest of this paper.

In addition, this definition of random walk kernel could be further generalized and enhanced by aggregating neighbor features non-linearly:
$$
\kernel_{loc}^n(v,v') = \inner{\f_v}{\f_{v'}} \circ \lambda \sum\limits_{u\in N(v)}\sum\limits_{u'\in N(v')} \kernel_{loc,\sigma}^{n-1}(u,u')
$$ 
where $\circ$ could be either multiplication or addition. $\sigma(\cdot)$ denotes a non-linear activation and $\kernel^{n-1}_{loc,\sigma}(\cdot,\cdot)$ denotes the post-activation kernel when $\sigma(\cdot)$ is involved. The generalized kernel could be realized by modifying Eq.(\ref{eq:rw_nn}) into:
\begin{equation}
\cc_j[v] = \W^{(j)} \f_v \circ \lambda \sum_{u\in N(v)} \sigma(c_{j-1}[u])
\end{equation}
where $\circ$ could be either $+$ or $\odot$ operation.

\subsection{Deep Graph Kernels and NNs}
Following Section~\ref{sec:string}, we could stack multiple graph kernel NNs to form a deep network. That is:
\begin{align}
\cc_1^{(l)}[v] &\ =\ \W^{(l,1)} \h^{(l-1)}[v] \nonumber \\
\cc_j^{(l)}[v] &\ =\ \W^{(l,j)} \h^{(l-1)}[v] \circ \lambda \sum_{u\in N(v)} \sigma\left(\cc_{j-1}^{(l)}[u]\right) \nonumber \\
\h^{(l)}[v] &\ =\ \sigma(\mbf{U}^{(l)}\cc_n^{(l)}[v]) \qquad\quad 1 \leq l\leq L,1<j\leq n\nonumber
\end{align}
The local kernel function is recursively defined in two dimensions: depth (term $\h^{(l)}$) and width (term $\cc_j$). Let the pre-activation kernel in the $l$-th layer be $\kernel^{(l)}_{loc}(v,v')=\kernel^{(l,n)}_{loc}(v,v')$, and the post-activation kernel be $\kernel^{(l)}_{loc,\sigma}(v,v')=\kernel^{(l,n)}_{loc,\sigma}(v,v')$. We recursively define $\kernel^{(l,j)}_{loc}(v,v')=$
$$
\begin{cases}
\displaystyle
\ \kernel^{(l-1)}_{loc,\sigma}(v,v') & \text{if } j=1 \\[0.5ex]
\ \kernel^{(l-1)}_{loc,\sigma}(v,v') \circ \lambda \sum\limits_{u\in N(v)}\sum\limits_{u'\in N(v')} \kernel^{(l,j-1)}_{loc,\sigma}(u,u') & \text{if } j>1
\end{cases}
$$
for $j=1,\cdots,n$. Finally, the graph kernel is $\kernel^{(L,n)}(G,G')=\sum_{v,v'}\kernel_{loc}^{(L,n)}(v,v')$. Similar to Theorem~\ref{theorem:deep_act}, we have
\begin{theo}
Consider a deep graph kernel NN with $L$ layers and activation function $\sigma(\cdot)$. Let the final output state $\h_G=\sum_v\h^{(L)}[v]$. For $l=1,\cdots,L;j=1,\cdots,n$:
\begin{enumerate}[itemsep=0pt]
\vspace{-5pt}
\item[(i)] $\mbf{c}^{(l)}_j[v][i]$ as a function of input $v$ and graph $G$ belongs to the RKHS of kernel $\kernel^{(l,j)}_{loc}(\cdot,\cdot)$;
\item[(ii)] $\h^{(l)}[v][i]$ belongs to the RKHS of kernel $\kernel^{(l,n)}_{loc,\sigma}(\cdot,\cdot)$.
\item[(iii)] $\h_G[i]$ belongs to the RKHS of kernel $\kernel^{(L,n)}(\cdot,\cdot)$.
\end{enumerate}
\label{theorem:deep_graph_nn}
\end{theo}

\subsection{Connection to Weisfeiler-Lehman Kernel}
We derived the above deep kernel NN for the purpose of generality. This model could be simplified by setting $n=2$, without losing representational power (as non-linearity is already involved in depth dimension). In this case, we rewrite the network by reparametrization:
\begin{equation}
\h^{(l)}_v = \sigma\left(\mbf{U}_1^{(l)}\h^{(l-1)}_v \circ \mbf{U}_2^{(l)}\sum_{u\in N(v)} \sigma\left(\mbf{V}^{(l)} \h^{(l-1)}_u\right)\right) \label{eq:wl-iterate}
\end{equation}
In this section, we further show that this model could be enhanced by sharing weight matrices $\mbf{U}$ and $\mbf{V}$ across layers. This parameter tying mechanism allows our model to embed Weisfeiler-Lehman kernel~\cite{shervashidze2011weisfeiler}. For clarity, we briefly review basic concepts of Weisfeiler-Lehman kernel below.

\begin{figure}
\includegraphics[scale=0.285]{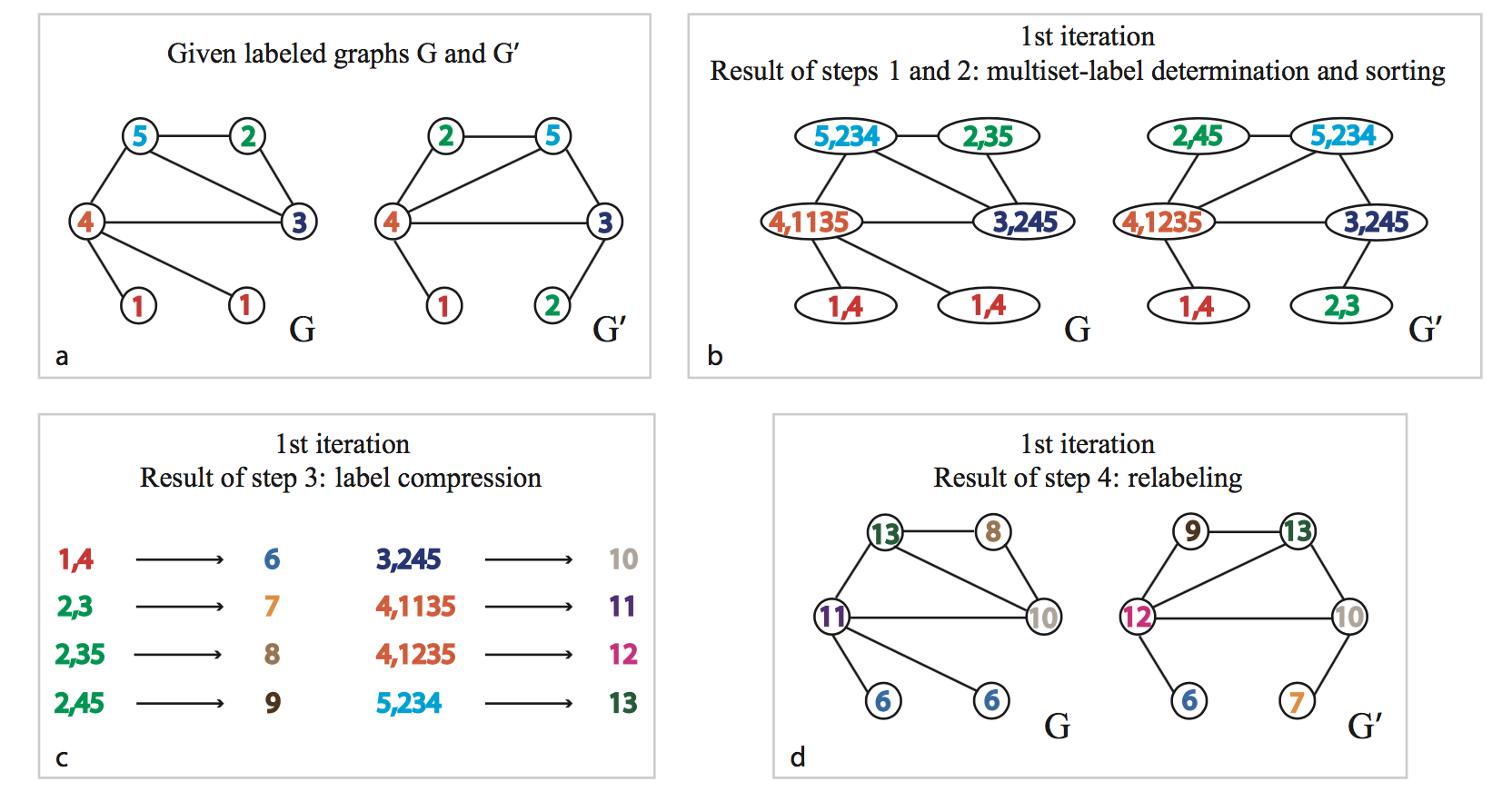}
\caption{Node relabeling in Weisfeiler-Lehman isomorphism test. Figure taken from \citet{shervashidze2011weisfeiler}}
\label{fig:wl-relabel}
\vspace{-10pt}
\end{figure}

{\bf Weisfeiler-Lehman Graph Relabeling } Weisfeiler-Lehman kernel borrows concepts from the Weisfeiler-Lehman isomorphism test for labeled graphs. The key idea of the algorithm is to augment the node labels by the sorted set of node labels of neighbor nodes, and compress these augmented labels into new, short labels (Figure \ref{fig:wl-relabel}). Such relabeling process is repeated $T$ times. In the $i$-th iteration, it generates a new labeling $l_i(v)$ for all nodes $v$ in graph $G$, with initial labeling $l_0$.

{\bf Generalized Graph Relabeling } The key observation here is that graph relabeling operation could be viewed as neighbor feature aggregation. As a result, the relabeling process naturally generalizes to the case where nodes are associated with continuous feature vectors. In particular, let $r$ be the relabeling function. For a node $v\in G$:
\begin{equation}
r(v) = \sigma(\mbf{U}_1 \f_v + \mbf{U}_2\sum_{u\in N(v)} \sigma(\mbf{V} \f_u))
\end{equation}
Note that our definition of $r(v)$ is exactly the same as $\h_v$ in Equation \ref{eq:wl-iterate}, with $\circ$ being additive composition.
\paragraph{Weisfeiler-Lehman Kernel} Let $\kernel$ be any graph kernel (called base kernel). Given a relabeling function $r$, Weisfeiler-Lehman kernel with base kernel $\kernel$ and depth $L$ is defined as
\begin{equation}
\kernel_{WL}^{(L)}(G,G') = \sum_{i=0}^L \kernel(r^i(G),r^i(G'))
\end{equation}
where $r^0(G)=G$ and $r^i(G),r^i(G')$ are the $i$-th relabeled graph of $G$ and $G'$ respectively. 

{\bf Weisfeiler-Lehman Kernel NN } Now with the above kernel definition, and random walk kernel as the base kernel, we propose the following recurrent module:
\begin{eqnarray}
\cc_0^{(l)}[v] &=& \W^{(l,0)} \h_v^{(l-1)} \nonumber \\
\cc_j^{(l)}[v] &=& \lambda\sum_{u\in N(v)} c_{j-1}^{(l)}[u] \odot \W^{(l,j)} \h_v^{(l-1)} \nonumber \\ 
\h_v^{(l)} &=& \sigma\left(\mbf{U}_1 \h_v^{(l-1)} + \mbf{U}_2\sum_{u\in N(v)} \sigma(\mbf{V} \h_u^{(l-1)})\right) \nonumber \\
\h_G^{(l)} &=& \sum_v\cc_n^{(l)}[v] \qquad\quad 1\leq l\leq L,1<j\leq n\nonumber
\end{eqnarray}
where $\h_v^{(0)}=\f_v$ and $\mbf{U}_1,\mbf{U}_2,\mbf{V}$ are shared across layers. The final output of this network is $\h_G = \sum_{l=1}^L \h_G^{(l)}$. 

The above recurrent module is still an instance of deep kernel, even though some parameters are shared. A minor difference here is that there is an additional random walk kernel NN that connects $i$-th layer and the output layer. But this is just a linear combination of $L$ deep random walk kernels (of different depth). Therefore, as an corollary of Theorem~\ref{theorem:deep_graph_nn}, we have:
\begin{prop}
For a Weisfeiler-Lehman Kernel NN with $L$ iterations and random walk kernel $\kernel_W^n$ as base kernel, the final output state $\h_G=\sum_l\h^{(l)}_G$ belongs to the RKHS of kernel $\kernel_{WL}^{(L)}(\cdot,\cdot)$.
\label{theorem:wl_nn}
\end{prop}

%% file: gated_rcnn.tex
\section{Adaptive Decay with Neural Gates}
\label{sec:gate}
The sequence and graph kernel (and their neural components) discussed so far use a constant decay value $\lambda$ regardless of the current input.
However, this is often not the case since the importance of the input can vary across the context or the applications. One extension is to make use of neural gates that adaptively control the decay factor. Here we give two illustrative examples:

{\bf Gated String Kernel NN }
By replacing constant decay $\lambda$ with a sigmoid gate, we modify our single-layer sequence module as:
\begin{align*}
 \lambda_t &= \sigma(\mbf{U}[\x_t, \h_{t-1}] + \mbf{b}) \\
 \mbf{c}_1[t] &= \lambda_t \odot \mbf{c}_1[t-1] + \left(\W^{(1)} \x_t\right)\\
 \mbf{c}_j[t] &= \lambda_t \odot \mbf{c}_j[t-1] + \left(\mbf{c}_{j-1}[t-1]\odot\W^{(j)}\x_t\right) \\
 \mbf{h}[t] &= \sigma(\mbf{c}_n[t]) \qquad\qquad\qquad\quad 1<j\leq n
\end{align*}
As compared with the original string kernel, now the decay factor $\lambda_{\x,i,j}$ is no longer $\lambda^{|\x|-i-1}$, but rather an adaptive value based on current context.

{\bf Gated Random Walk Kernel NN } Similarly, we could introduce gates so that different walks have different weights:
\begin{eqnarray}
\lambda_{u,v} &=& \sigma(\mbf{U} [\f_u,\f_v] + \mbf{b}) \nonumber \\
\cc_0[v] &=& \W^{(0)} \f_v \nonumber \\
\cc_j[v] &=& \sum_{u\in N(v)} \lambda_{u,v} \odot c_{j-1}[u] \odot \W^{(j)} \f_v \nonumber \\ 
\h_G &=& \sigma(\sum_v\cc_n[v]) \qquad\quad 1<j\leq n\nonumber
\end{eqnarray}
The underlying kernel of the above gated network becomes
\begin{equation}
\kernel_W^n(G,G') = \sum_{\x \in P_n(G)} \sum_{\y \in P_n(G')} \prod_{i=1}^n \lambda_{\x_i,\y_i}\inner{\f_{\x_i}}{\f_{\y_i}} \nonumber
\end{equation}
where each path is weighted by different decay weights, determined by network itself.

%% file: rel_work.tex
\section{Related Work}
{\bf Sequence Networks } Considerable effort has gone into designing effective networks for sequence processing. This includes recurrent modules with the ability to carry persistent memories such as LSTM~\cite{hochreiter1997long} and GRU~\cite{chung2014empirical}, as well as non-consecutive convolutional modules (RCNNs, \citet{lei2015semi}), and others. More recently, \citet{zoph2016neural} exemplified a reinforcement learning-based search algorithm to further optimize the design of such recurrent architectures. Our proposed neural networks offer similar state evolution and feature aggregation functionalities but derive the motivation for the operations involved from well-established kernel computations over sequences. 

Recursive neural networks are alternative architectures to model hierarchical structures such as syntax trees and logic forms.
For instance, \citet{socher2013sentiment} employs recursive networks for sentence classification, where each node in the dependency tree of the sentence is transformed into a vector representation. \citet{tai2015improved} further proposed tree-LSTM, which incorporates LSTM-style architectures as the transformation unit.
\citet{dyer-EtAl:2015,dyer-EtAl:2016} recently introduced a recursive neural model for transition-based language modeling and parsing.
While not specifically discussed in the paper, our ideas do extend to similar neural components for hierarchical objects (e.g. trees).

{\bf Graph Networks } Most of the current graph neural architectures perform either convolutional or recurrent operations on graphs. \citet{duvenaud2015convolutional} developed Neural Fingerprint for chemical compounds, where each convolution operation is a sum of neighbor node features, followed by a linear transformation. Our model differs from theirs in that our generalized kernels and networks can aggregate neighboring features in a non-linear way. Other approaches, e.g., \citet{bruna2013spectral} and \citet{henaff2015deep}, rely on graph Laplacian or Fourier transform. 

For recurrent architectures, \citet{li2015gated} proposed gated graph neural networks, where neighbor features are aggregated by GRU function. \citet{dai2016discriminative} considers a different architecture where a graph is viewed as a latent variable graphical model. Their recurrent model is derived from Belief Propagation-like algorithms. Our approach is most closely related to \citet{dai2016discriminative}, in terms of neighbor feature aggregation and resulting recurrent architecture. Nonetheless, the focus of this paper is on providing a framework for how such recurrent networks could be derived from deep graph kernels.

{\bf Kernels and Neural Nets } Our work follows recent work demonstrating the connection between neural networks and kernels~\cite{cho2009kernel,hazan2015steps}.
For example, \citet{zhang16l1} showed that standard feedforward neural nets belong to a larger space of recursively constructed kernels (given certain activation functions).
Similar results have been made for convolutional neural nets~\cite{anselmi2015deep}, and general computational graphs~\cite{DanielyFS16}.
We extend prior work to kernels and neural architectures over structured inputs, in particular, sequences and graphs.
Another difference is how we train the model. While some prior work appeals to convex optimization through improper learning~\cite{zhang16l1,heinemann2016improper} (since kernel space is larger), we use the proposed networks as building blocks in typical non-convex but flexible neural network training.

%% file: experiment.tex
\section{Experiments}
\label{sec:expr}
The left-over question is whether the proposed class of operations, despite its formal characteristics, leads to more effective architecture exploration and hence improved performance.
In this section, we apply the proposed sequence and graph modules to various tasks and empirically evaluate their performance against other neural network models. These tasks include language modeling, sentiment classification and molecule regression.

\subsection{Language Modeling on PTB}

{\bf Dataset and Setup } 
We use the Penn Tree Bank (PTB) corpus as the benchmark. The dataset contains about 1 million tokens in total.
We use the standard train/development/test split of this dataset with vocabulary of size 10,000. 

{\bf Model Configuration } 
Following standard practice, we use SGD with an initial learning rate of 1.0 and decrease the learning rate by a constant factor after a certain epoch. We back-propagate the gradient with an unroll size of 35 and use dropout~\cite{dropout} as the regularization.
Unless otherwise specified, we train 3-layer networks with $n=1$ and normalized adaptive decay.\footnote{See the supplementary sections for a discussion of network variants.}
Following~\cite{zilly2016recurrent}, we add highway connections~\cite{srivastava2015training} within each layer:
\begin{align*}
\cc^{(l)}[t] &= \lambda_t\odot \cc^{(l)}[t-1]+(1-\lambda_t)\odot(\W^{(l)}\h^{(l-1)}[t])\\
\h^{(l)}[t] &= \mathbf{f}_t\odot\cc^{(l)}[t] + (1-\mathbf{f}_t)\odot \h^{(l-1)}[t]
\end{align*}
where $\h^{(0)}[t]=\x_t$, $\lambda_t$ is the gated decay factor and $\mathbf{f}_t$ is the transformation gate of highway connections.\footnote{We found non-linear activation is no longer necessary when the highway connection is added.}

\newcommand\Tstrut{\rule{0pt}{2.0ex}}         
\newcommand\Bstrut{\rule[-0.9ex]{0pt}{0pt}}   
\begin{table}[t!]
\centering
\caption{Comparison with state-of-the-art results on PTB. $\vert\theta\vert$ denotes the number of parameters. Following recent work~\cite{press2016using}, we share the input and output word embedding matrix.
We report the test perplexity (PPL) of each model. Lower number is better.
}
\label{table:lm_stoa}
\begin{tabular}{lcc}
\hline
\textbf{Model} & $\vert\theta\vert$ & {\bf PPL} \Tstrut\Bstrut\\
\hline
LSTM (large)~\cite{zaremba2014recurrent} & 66m & 78.4 \Tstrut\Bstrut\\
Character CNN~\cite{kim2015character} & 19m & 78.9 \Tstrut\Bstrut\\
Variational LSTM~(\citeauthor{Gal2016Theoretically}) & 20m & 78.6 \Tstrut\Bstrut\\
Variational LSTM~(\citeauthor{Gal2016Theoretically}) & 66m & 73.4 \Tstrut\Bstrut\\
Pointer Sentinel-LSTM~(\citeauthor{merity2016pointer}) & 21m & 70.9 \Tstrut\Bstrut\\
Variational RHN~\cite{zilly2016recurrent} & 23m & 65.4 \Tstrut\Bstrut\\
Neural Net Search~\cite{zoph2016neural} & 25m & 64.0 \Tstrut\Bstrut\\
\hline
Kernel NN ($\lambda = 0.8$) & 5m & 84.3 \Tstrut\Bstrut\\
Kernel NN ($\lambda$ learned as parameter) & 5m & 76.8 \Tstrut\Bstrut\\
\hline
Kernel NN (gated $\lambda$) & 5m & 73.6 \Tstrut\Bstrut\\
Kernel NN (gated $\lambda$) & 20m  & 69.2 \Tstrut\Bstrut\\
$\ \ $ + variational dropout  & 20m & 65.5 \Tstrut\Bstrut\\
$\ \ $ + variational dropout, 4 RNN layers  & 20m & {\bf 63.8} \Tstrut\Bstrut\\
\hline
\end{tabular}
\end{table}
{\bf Results } Table~\ref{table:lm_stoa} compares our model with various state-of-the-art models. 
Our small model with 5 million parameters achieves a test perplexity of 73.6, already outperforming many results achieved using much larger network.
By increasing the network size to 20 million, we obtain a test perplexity of 69.2, with standard dropout.
Adding variational dropout~\cite{Gal2016Theoretically} within the recurrent cells further improves the perplexity to 65.5. 
Finally, the model achieves 63.8 perplexity when the recurrence depth is increased to 4, being state-of-the-art and on par with the results reported in~\cite{zilly2016recurrent,zoph2016neural}.
Note that \citet{zilly2016recurrent} uses 10 neural layers and \citet{zoph2016neural} adopts a complex recurrent cell found by reinforcement learning based search.
Our network is architecturally much simpler.

\begin{figure}
\centering
\vspace{0.05in}
\includegraphics[width=2.8in]{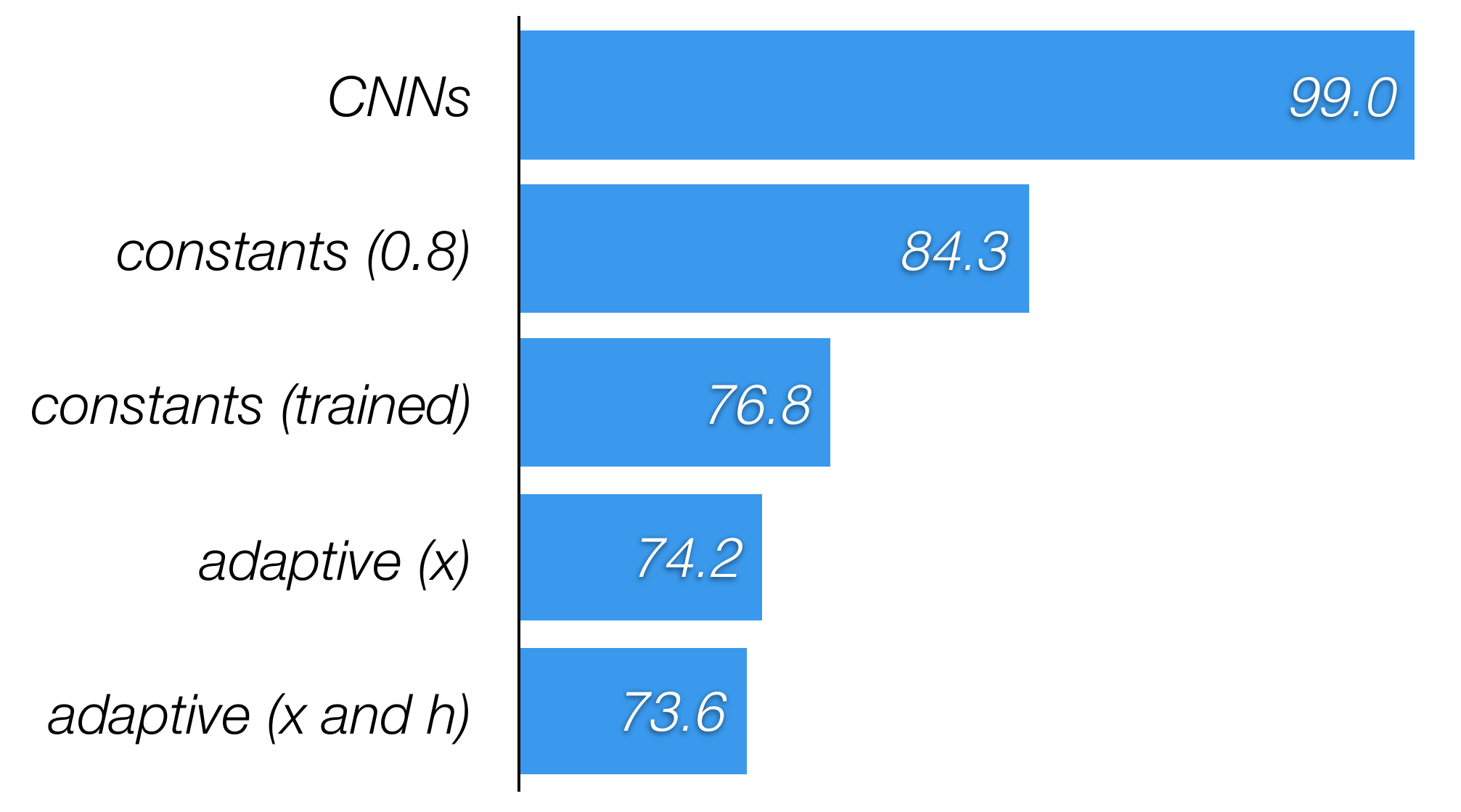}
\vspace{-0.1in}
\caption{Comparison between kernel NN variants on PTB. $|\theta|=5\text{m}$ for all models. Hyper-parameter search is performed for each variant.}
\label{fig:lm-ana}
\vspace{-0.15in}
\end{figure}

Figure~\ref{fig:lm-ana} analyzes several variants of our model. 
Word-level CNNs are degraded cases ($\lambda=0$) that ignore non-contiguous n-gram patterns.
Clearly, this variant performs worse compared to other recurrent variants with $\lambda >0$. Moreover, the test perplexity improves from 84.3 to 76.8 when we train the constant decay vector as part of the model parameters.
Finally, the last two variants utilize neural gates (depending on input $\x_t$ only or both input $\x_t$ and previous state $\h[t-1]$), further improving the performance.

\subsection{Sentiment Classification}
{\bf Dataset and Setup } We evaluate our model on the sentence classification task. We use the Stanford Sentiment Treebank benchmark~\cite{socher2013sentiment}. The dataset consists of 11855 parsed English sentences annotated at both the root (i.e. sentence) level and the phrase level using 5-class fine-grained labels. We use the standard split for training, development and testing. Following previous work, we also evaluate our model on the binary classification variant of this benchmark, ignoring all neutral sentences. 

Following the recent work of DAN~\cite{iyyer2015} and RLSTM~\cite{tai2015improved}, we use the publicly available 300-dimensional GloVe word vectors~\cite{pennington2014glove}.
Unlike prior work which fine tunes the word vectors, we normalize the vectors (i.e. $\|w\|_2^2=1$) and fixed them for simplicity.

{\bf Model Configuration } Our best model is a 3-layer network with $n=2$ and hidden dimension $d = 200$. We average the hidden states $h[t]$ across $t$, and concatenate the averaged vectors from the 3 layers as the input of the final softmax layer.
The model is optimized with Adam \cite{kingma2015adam}, and dropout probability of 0.35.

\begin{table}[t!]
\centering
\caption{Classification accuracy on Stanford Sentiment Treebank. 
Block I: recursive networks; Block II: convolutional or recurrent networks; Block III: other baseline methods. Higher number is better.
}
\vspace{-3pt}
\label{table:sentiment}
\begin{tabular}{l@{~~~~~~~}c@{~~~~~}c@{~~~}}
\hline
{\bf Model} & {\bf Fine} & {\bf Binary} \Tstrut\Bstrut\\
\hline
RNN~(\citet{socher2011semi}) & 43.2 & 82.4 \Tstrut\Bstrut\\
RNTN~(\citet{socher2013sentiment}) & 45.7 & 85.4 \Tstrut\Bstrut\\
DRNN~(\citet{irsoy2014}) & 49.8 & 86.8 \Tstrut\Bstrut\\
RLSTM~(\citet{tai2015improved}) & 51.0 & 88.0 \Tstrut\Bstrut\\
\hline
DCNN~(\citet{kalchbrenner2014}) & 48.5 & 86.9 \Tstrut\Bstrut\\
CNN-MC~(\citet{Kim14}) & 47.4 & 88.1 \Tstrut\Bstrut\\
Bi-LSTM~(\citet{tai2015improved}) & 49.1 & 87.5 \Tstrut\Bstrut\\
LSTMN~(\citet{cheng2016}) & 47.9 & 87.0 \Tstrut\Bstrut\\
\hline
PVEC~(\citet{le2014distributed}) & 48.7 & 87.8 \Tstrut\Bstrut\\
DAN~(\citet{iyyer2014}) & 48.2 & 86.8 \Tstrut\Bstrut\\
DMN~(\citet{KumarISBEPOGS15}) & 52.1 & 88.6 \Tstrut\Bstrut\\
\hline
Kernel NN, $\lambda=0.5$ & 51.2 & 88.6 \Tstrut\Bstrut\\
Kernel NN, gated $\lambda$ & \textbf{53.2} & \textbf{89.9} \Tstrut\Bstrut\\
\hline
\end{tabular}
\vspace{-6pt}
\end{table}

{\bf Results }
Table~\ref{table:sentiment} presents the performance of our model and other networks. 
We report the best results achieved across 5 independent runs.
Our best model obtains 53.2\% and 89.9\% test accuracies on fine-grained and binary tasks respectively.
Our model with only a constant decay factor also obtains quite high accuracy, outperforming other baseline methods shown in the table.

\begin{table}[!t!]
\centering
\caption{Experiments on Harvard Clean Energy Project. We report Root Mean Square Error(RMSE) on test set. The first block lists the results reported in \citet{dai2016discriminative} for reference. For fair comparison, we reimplemented their best model so that all models are trained under the same setup. Results under our setup is reported in second block.}
\vspace{-3pt}
\label{tab:cep}
\begin{tabular}{lcc}
\hline
\textbf{Model} \cite{dai2016discriminative} & $\vert\theta\vert$ & {\bf RMSE} \Tstrut\Bstrut\\
\hline
Mean Predicator & 1 & 2.4062 \Tstrut\Bstrut\\
Weisfeiler-lehman Kernel, degree=3 & 1.6m & 0.2040 \Tstrut\Bstrut\\
Weisfeiler-lehman Kernel, degree=6 & 1378m & 0.1367 \Tstrut\Bstrut\\
Embedded Mean Field & 0.1m & 0.1250 \Tstrut\Bstrut\\
Embedded Loopy BP & 0.1m & 0.1174 \Tstrut\Bstrut\\
\hline
\textbf{Under Our Setup} & & \Tstrut\Bstrut \\
\hline
Neural Fingerprint & 0.26m & 0.1409 \Tstrut\Bstrut\\
Embedded Loopy BP & 0.26m & 0.1065 \Tstrut\Bstrut\\
\hline
Weisfeiler Kernel NN  & 0.26m & 0.1058 \Tstrut\Bstrut\\
Weisfeiler Kernel NN, gated $\lambda$  & 0.26m & \textbf{0.1043}  \Tstrut\Bstrut\\
\hline
\end{tabular}
\vspace{-8pt}
\end{table}

\subsection{Molecular Graph Regression}
{\bf Dataset and Setup } We further evaluate our graph NN models on the Harvard Clean Energy Project benchmark, which has been used in \citet{dai2016discriminative,duvenaud2015convolutional} as their evaluation dataset. This dataset contains 2.3 million candidate molecules, with each molecule labeled with its power conversion efficiency (PCE) value.

We follow exactly the same train-test split as~\citet{dai2016discriminative}, and the same re-sampling procedure on the training data (but not the test data) to make the algorithm put more emphasis on molecules with higher PCE values, since the data is distributed unevenly. 

We use the same feature set as in~\citet{duvenaud2015convolutional} for atoms and bonds. Initial atom features include the atom’s element, its degree, the number of attached hydrogens, its implicit valence, and an aromaticity indicator. The bond feature is a concatenation of bond type indicator, whether the bond is conjugated, and whether the bond is in a ring.

{\bf Model Configuration } Our model is a Weisfeiler-Lehman NN, with 4 recurrent iterations and $n=2$. All models (including baseline) are optimized with Adam \cite{kingma2015adam}, with learning rate decay factor 0.9. 

{\bf Results } In Table \ref{tab:cep}, we report the performance of our model against other baseline methods. Neural Fingerprint~\cite{duvenaud2015convolutional} is a 4-layer convolutional neural network. Convolution is applied to each atom, which sums over its neighbors' hidden state, followed by a linear transformation and non-linear activation. Embedded Loopy BP~\cite{dai2016discriminative} is a recurrent architecture, with 4 recurrent iterations. It maintains message vectors for each atom and bond, and propagates those vectors in a message passing fashion. Table \ref{tab:cep} shows our model achieves state-of-the-art against various baselines.

%% file: conclusion.tex
\section{Conclusion}
We proposed a class of deep recurrent neural architectures and formally characterized its underlying computation using kernels.
By linking kernel and neural operations, we have a ``template'' for deriving new families of neural architectures for sequences and graphs.
We hope the theoretical view of kernel neural networks can be helpful for future model exploration.

\section*{Acknowledgement}
We thank Prof. Le Song for sharing Harvard Clean Energy Project dataset. We also thank Yu Zhang, Vikas Garg, David Alvarez, Tianxiao Shen, Karthik Narasimhan and the reviewers for their helpful comments. This work was supported by the DARPA Make-It program under contract ARO W911NF-16-2-0023.

%% file: appendix.tex
\onecolumn
\appendix
\pagestyle{empty}
\newgeometry{left=3.5cm,right=3.5cm}
\icmltitle{Supplementary Material}
\section{Examples of kernel / neural variants}
Our theoretical results apply to some other variants of sequence kernels and the associated neural components. We give some examples in the this section. Table~\ref{table:rcnn_dp} shows three network variants, corresponding to three realizations of string kernels provided in Table~\ref{table:kernel_summarization}. 

\paragraph{Connection to LSTMs}
Interestingly, many recent work has reached similar RNN architectures through empirical exploration. For example, \citet{greff2015lstm} found that simplifying LSTMs, by removing the input gate or coupling it with the forget gate does not significantly change the performance.
However, the forget gate (corresponding to the decay factor $\lambda$ in our notation) is crucial for performance.
This is consistent with our theoretical analysis and the empirical results in Figure~\ref{fig:lm-ana}.
Moreover, \citet{BalduzziG16} and \citet{ran17} both suggest that a linear additive state computation suffices to provide competitive performance compared to LSTMs:
\footnote{\citet{BalduzziG16} also includes the previous token , i.e. $\W \x_t + \W' \x_{t-1}$, which doesn't affect the discussion here.}
\begin{align*}
 \mbf{c}[t] &= \lambda_f \odot \mbf{c}_1[t-1] + \lambda_i\odot \left(\W \x_t\right)\\
 \mbf{h}[t] &= \sigma(\mbf{c}[t])
\end{align*}
In fact, this variant becomes an instance of the kernel NN presented in this work (with $n=1$ and adaptive gating), when $\lambda_f=\lambda_t$ and $\lambda_i=1-\lambda_t$ or 1.

\vspace{0.3in}
\begin{table}[!h!]
\small
\centering
\begin{tabular}{|@{~~~~~~~~}l@{~~~~}l@{~~~~~~~~~~~~~~~~~~~~~~}|}
\hline
 &  \\
 \multicolumn{2}{|l|}{\bf ~~(a) Multiplicative mapping, aggregation un-normalized:} \\[12pt]
& $\displaystyle \mbf{c}_1[t] = \lambda\cdot\mbf{c}_1[t-1] + \W^{(1)}\x_t $ \\[6pt]
& $\displaystyle \mbf{c}_2[t] = \lambda\cdot\mbf{c}_2[t-1] + \left(\mbf{c}_1[t-1] \odot \W^{(2)}\x_t \right) $ \\[6pt]
& $\displaystyle \mbf{c}_3[t] = \lambda\cdot\mbf{c}_3[t-1] + \left(\mbf{c}_2[t-1] \odot \W^{(3)}\x_t \right) $ \\[6pt]
 &  \\
\hline
 &  \\
 \multicolumn{2}{|l|}{\bf ~~(b) Multiplicative mapping, aggregation \underline{normalized}:} \\[12pt]
& $\displaystyle \mbf{c}_1[t] = \lambda\cdot\mbf{c}_1[t-1] + (1-\lambda)\cdot \W^{(1)}\x_t $ \\[6pt]
& $\displaystyle \mbf{c}_2[t] = \lambda\cdot\mbf{c}_2[t-1] + (1-\lambda)\cdot\left(\mbf{c}_1[t-1] \odot \W^{(2)}\x_t \right) $ \\[6pt]
& $\displaystyle \mbf{c}_3[t] = \lambda\cdot\mbf{c}_3[t-1] + (1-\lambda)\cdot\left(\mbf{c}_2[t-1] \odot \W^{(3)}\x_t \right) $ \\[6pt]
 &  \\
\hline
 &  \\
 \multicolumn{2}{|l|}{\bf ~~(c) \underline{Additive} mapping, aggregation normalized:} \\[12pt]
& $\displaystyle \mbf{c}_1[t] = \lambda\cdot\mbf{c}_1[t-1] + (1-\lambda)\cdot \W^{(1)}\x_t $ \\[6pt]
& $\displaystyle \mbf{c}_2[t] = \lambda\cdot\mbf{c}_2[t-1] + (1-\lambda)\cdot\left(\mbf{c}_1[t-1] + \W^{(2)}\x_t \right) $ \\[6pt]
& $\displaystyle \mbf{c}_3[t] = \lambda\cdot\mbf{c}_3[t-1] + (1-\lambda)\cdot\left(\mbf{c}_2[t-1] + \W^{(3)}\x_t \right) $ \\[6pt]
 &  \\
\hline
&  \\
\multicolumn{2}{|l|}{\bf ~~Final activation:} \\[12pt]
& $\displaystyle \h[t] = \sigma\left(\mbf{c}_3[t]\right) $ \\[6pt]
& $\displaystyle \h[t] = \sigma\left(\mbf{c}_1[t]+\mbf{c}_2[t]+\mbf{c}_3[t]\right) $ $\qquad$ (any linear combination of $\mbf{c}_*[t]$) \\[6pt]
&  \\
\hline
\end{tabular}
\caption{Example sequence NN variants. We present these equations in the context of $n=3$.}
\label{table:rcnn_dp}
\end{table}

\begin{table}[!h!]
\small
\centering
\begin{tabular}{|@{~~~~~~~~~}lll@{~~~~~~~~~~}|}
\hline
 & & \\
 \multicolumn{3}{|l|}{\bf ~~(a) Multiplicative mapping, aggregation un-normalized:} \\[15pt]
& $\displaystyle \kernel_2(\x,\y) = \sum_{1\leq i<j\leq\len{\x}} \sum_{1\leq k<l\leq\len{\y}} \lambda^{\len{\x}-i-1}\,\lambda^{\len{\y}-k-1} \ \inner{\x_i}{\y_k}\cdot\inner{\x_j}{\y_l}$ & \\[24pt]
& $\displaystyle \quad\, \phi_2(\x) = \sum_{1\leq i<j\leq\len{\x}} \lambda^{\len{\x}-i-1}\ \x_i\otimes\x_j$ & \\
 & & \\
\hline
 & & \\
 \multicolumn{3}{|l|}{\bf ~~(b) Multiplicative mapping, aggregation \underline{normalized}:} \\[15pt]
 & $\displaystyle \kernel_2(\x,\y) =\ \frac{1}{Z} \sum_{1\leq i<j\leq\len{\x}} \sum_{1\leq k<l\leq\len{\y}} \lambda^{\len{\x}-i-1}\,\lambda^{\len{\y}-k-1}\ \inner{\x_i}{\y_k}\cdot\inner{\x_j}{\y_l}$ & \\[24pt]
& $\qquad $ s.t.$\quad$ $\displaystyle Z = \sum_{1\leq i<j\leq\len{\x}} \sum_{1\leq k<l\leq\len{\y}} \lambda^{\len{\x}-i-1}\,\lambda^{\len{\y}-k-1}$ & \\[30pt]
& $\displaystyle \quad\, \phi_2(\x) =\ \frac{1}{Z'} \sum_{1\leq i<j\leq\len{\x}} \lambda^{\len{\x}-i-1}\ \x_i\otimes\x_j$ & \\[24pt]
& $\qquad$ s.t.$\quad$ $\displaystyle Z' = \sum_{1\leq i<j\leq\len{\x}} \lambda^{\len{\x}-i-1}$ & \\
 & & \\
\hline
 & & \\
 \multicolumn{3}{|l|}{\bf ~~(c) \underline{Additive} mapping, aggregation normalized:} \\[15pt]
  & $\displaystyle \kernel_2(\x,\y) =\ \frac{1}{Z} \sum_{1\leq i<j\leq\len{\x}} \sum_{1\leq k<l\leq\len{\y}} \lambda^{\len{\x}-i-1}\,\lambda^{\len{\y}-k-1}\ \left(\inner{\x_i}{\y_k}+\inner{\x_j}{\y_l}\right)$ & \\[24pt]
& $\qquad $ s.t.$\quad$ $\displaystyle Z = \sum_{1\leq i<j\leq\len{\x}} \sum_{1\leq k<l\leq\len{\y}} \lambda^{\len{\x}-i-1}\,\lambda^{\len{\y}-k-1}$ & \\[30pt]
& $\displaystyle \quad\, \phi_2(\x) =\ \frac{1}{Z'} \sum_{1\leq i<j\leq\len{\x}} \lambda^{\len{\x}-i-1}\ [\x_i,\x_j]$ & \\[24pt]
& $\qquad$ s.t.$\quad$ $\displaystyle Z' = \sum_{1\leq i<j\leq\len{\x}} \lambda^{\len{\x}-i-1}$ & \\
 & & \\
\hline
\end{tabular}
\caption{Examples of sequence kernel functions and associated mappings. $[\x_i,\x_j]$ denotes the concatenation of two vectors.}
\label{table:kernel_summarization}
\end{table}

\clearpage
\section{Proof of Theorem~\ref{theorem:onelayer}}
\label{appendix:onelayer_proof}


We first generalize the kernel definition in Eq.(\ref{eq:seqkernel-mul}) to the case of any n-gram. For any integer $n>0$, the underlying mapping of the $n$-th order string kernel is defined as,
\begin{align*}
\phi_n(\x)\ &=\ \sum_{1\leq i_1<\cdots <i_n\leq \len{\x}} \lambda^{\len{x}-i_1-n+1}\,\x_{i_1}\otimes\x_{i_2}\otimes\cdots\otimes\x_{i_n}
\end{align*}
We now show that String Kernel NN states $\cc_n[t]$ defined in~(\ref{eq:seqnn}),
\begin{align*}
n=1:
\qquad\qquad \mbf{c}_1[t] &= \lambda\cdot\mbf{c}_1[t-1] + \W^{(1)}\x_t \\
n>1:
\qquad\qquad \mbf{c}_n[t] &= \lambda\cdot\mbf{c}_n[t-1] + \left(\mbf{c}_{n-1}[t-1]\odot\W^{(n)}\x_t\right)
\end{align*}
is equivalent to summing over all n-grams within the first $t$ tokens $\x_{1:t}=\{\x_1,\cdots,\x_t\}$,
\begin{align*}
\mbf{c}_n[t] &= \sum_{1\leq i_1<i_2<\cdots<i_n\leq t} \left( \W^{(1)}\x_{i_1}\odot\cdots\odot\W^{(n)}\x_{i_n} \right)\cdot\lambda^{t-i_1-n+1} \end{align*}
\paragraph{Proof:} We prove the equivalence by induction. When $n=1$, for $\mbf{c}_1[t]$ we have,
\begin{align*}
\mbf{c}_1[t]\ &=\ \ \ \sum_{1\leq i_1\leq t} \left(\W^{(1)}\x_{i_1}\right) \cdot\lambda^{t-i_1} \\
  \ &=\  \underbrace{\left(\sum_{1\leq i_1 \leq t-1} \left(\W^{(1)}\x_{i_1}\right) \cdot \lambda^{t-1-i_1}\right)}_{\mbf{c}_1[t-1]} \cdot \lambda \ + \ \W^{(1)}\x_t \\[6pt]
 \  &=\  \mbf{c}_1[t-1] \cdot \lambda \ + \ \W^{(1)}\x_t
\end{align*}
When $n=k>1$, suppose the state iteration holds for $1,\cdots, k-1$, we have,
\begin{align*}
\mbf{c}_n[t]\ &=\ \sum_{1\leq i_1<i_2<\cdots<i_n\leq t} \left( \W^{(1)}\x_{i_1}\odot\cdots\odot\W^{(n)}\x_{i_n} \right)\cdot\lambda^{t-i_1-n+1} \\
 &=\ \ \underbrace{\left(\sum_{1\leq i_1<i_2<\cdots<i_n\leq t-1} \left( \W^{(1)}\x_{i_1}\odot\cdots\odot\W^{(n)}\x_{i_n} \right)\cdot\lambda^{t-1-i_1-n+1}\right)}_{\text{when } i_n<t:\ \ \mbf{c}_{n}[t-1]} \cdot\lambda \\
 &\ \ +\, \underbrace{\left(\sum_{1\leq i_1<i_2<\cdots<i_n=t} \left( \W^{(1)}\x_{i_1}\odot\cdots\odot\W^{(n)}\x_{i_n} \right)\cdot\lambda^{t-1-i_1-n+1}\right)}_{\text{when } i_n=t} \\
 &=\ c_{n}[t-1]\cdot\lambda \ + \ \left(\mbf{c}_{n-1}[t-1]\odot\W^{(n)}\x_t\right) \qquad \qed 
\end{align*}

Now we proceed to prove Theorem~\ref{theorem:onelayer}. 

\paragraph{Proof: } 
From the above results, we know that for $\mbf{c}_n[t]$,
\begin{align*}
\mbf{c}_n[t]\ &=\ \sum_{1\leq i_1<i_2<\cdots<i_n\leq t} \left( \W^{(1)}\x_{i_1}\odot\cdots\odot\W^{(n)}\x_{i_n} \right)\cdot\lambda^{t-i_1-n+1}
\end{align*}
In particular, the value of the i-th entry, $\mbf{c}_n[t][i]$, is equal to,
\begin{align}
\mbf{c}_n[t][i]\ &=\ \sum_{1\leq i_1<i_2<\cdots<i_n\leq t} \ 
\underbrace{\inner{\mbf{w}^{(1)}_i}{\x_{i_1}}\cdots\inner{\mbf{w}^{(n)}_i}{\x_{i_n}}}_{\inner{\x_{i_1}\otimes\cdots\otimes\x_{i_n}}{\mbf{w}^{(1)}_i\otimes\cdots\otimes\mbf{w}^{(n)}_i}}\ \cdot\ \lambda^{t-i_1-n+1}  \nonumber \\[6pt]
 &=\ \inner{\sum_{1\leq i_1<i_2<\cdots<i_n\leq t} \lambda^{t-i_1-n+1}\, \x_{i_1}\otimes\cdots\otimes\x_{i_n}}{\ \mbf{w}^{(1)}_i\otimes\cdots\otimes\mbf{w}^{(n)}_i} \nonumber \\[6pt]
 &= \ \inner{\phi_n(\x_{1:t})}{\phi_n(\mbf{w}_{i,n})} \nonumber
\end{align}
where $\mbf{w}^{(k)}_i$ represents the i-th row of matrix $\W^{(k)}$ and $\mbf{w}_{i,n}=\{\mbf{w}^{(1)}_i,\cdots,\mbf{w}^{(n)}_i\}$.
This completes the proof since $\kernel_n(\x_{1:t},\mbf{w}_{i,n})=\inner{\phi_n(\x_{1:t})}{\phi_n(\mbf{w}_{i,n})}$ by definition. \qed

\paragraph{Remarks} This theorem demonstrates that the NN state vectors $\mbf{c}_n$ can be seen as projecting the kernel mapping $\phi_n(\x)$ into a low-dimensional space using the parameterized matrices $\{\W^{(i)}\}_i=1^n$. 
This is similar to word embeddings and CNNs, where the input (a single word or an n-gram) is projected to obtain a low-dimensional representation.

\clearpage

\section{Proof of Theorem~\ref{theorem:deep_act}}
\label{appendix:deep_proof}

We first review necessary concepts and notations for the ease of reading. 
Similar to the proof in Appendix~\ref{appendix:onelayer_proof}, the generalized string kernel $\kernel^{(l)}$ and $\kernel_\sigma^{(l)}$ in Eq.(\ref{eq:sk_generalized}) can be defined with the underlying mappings,
\begin{align*}
\phi^{(l)}(\x)\ &=\ \sum_{1\leq i_1<\cdots <i_n\leq \len{\x}} \lambda^{\len{x}-i_1-n+1}\, \phi_\sigma^{(l-1)}(\x_{1:i_1}) \otimes \phi_\sigma^{(l-1)}(\x_{1:i_2}) \otimes \cdots \otimes \phi_\sigma^{(l-1)}(\x_{1:i_1}) \\[6pt]
\phi_\sigma^{(l)}(\x) \ &= \ \phi_\sigma(\phi^{(l)}(\x))
\end{align*}
where $\phi_\sigma()$ is the underlying mapping of a kernel function whose reproducing kernel Hilbert space (RKHS) contains the non-linear activation $\sigma()$ used in the String Kernel NN layer.
Here $\kernel^{(l)}()$ is the ``pre-activation kernel'' and $\kernel_\sigma^{(l)}()$ is the ``post-activation kernel''. To show that the values of String Kernel NN states $\mbf{c}^{(l)}[t]$ is contained in the RKHS of $\kernel^{(l)}()$ and that of $\h^{(l)}[t]$ is contained in the RKHS of $\kernel_\sigma^{(l)}()$, we re-state the claim in the following way,
\begin{theo}
Given a deep n-gram String Kernel NN with non-linear activation $\sigma()$. Assuming $\sigma()$ lies in the RKHS of a kernel function with underlying mapping $\phi_\sigma()$, then 
\begin{enumerate}
\item[(i)] $\mbf{c}^{(l)}[t]$ lies in the RKHS of kernel $\kernel^{(l)}()$ in the sense that
	\begin{align*}
    	\mbf{c}_j^{(l)}[t][i]\ =\ \inner{\phi^{(l)}(\x_{1:t})}{\mbf{\psi}^{(l)}_{i,j}}
    \end{align*}
for any internal state $\mbf{c}^{(l)}_j[t]$ ($1\leq j\leq n$) of the $l$-th layer, where $\psi^{(l)}_{i,j}$ is a mapping constructed from model parameters;
\item[(ii)] $\h^{(l)}[t]$ lies in the RKHS of kernel $\kernel_\sigma^{(l)}()$ as a corollary in the sense that
	\begin{align*}
    \h^{(l)}[t][i]\ &=\ \sigma(\mbf{c}_n^{(l)}[t][i]) \\
    	&=\ \sigma\left(\inner{\phi^{(l)}(\x_{1:t})}{\psi^{(l)}_{i,n}}\right) & (\text{based on }(i)) \\
        &=\ \inner{\phi_\sigma(\phi^{(l)}(\x_{1:t}))}{\psi_\sigma(\psi^{(l)}_{i,n})} & (\text{Lemma~\ref{lemma:zhang}})
    \end{align*}
and we denote $\psi_\sigma(\psi^{(l)}_{i,n})$ as  $\psi^{(l)}_{\sigma,i,n}$ for short. 
\end{enumerate}
\end{theo}
\paragraph{Proof:} 
We prove by induction on $l$. When $l=1$, the proof of Theorem~\ref{theorem:onelayer} already shows that $\mbf{c}^{(1)}_j[t][i]=\inner{\phi^{(1)}_j(\x_{1:t})}{\phi^{(1)}_j(\mbf{w}_{i,j})}$ in a one-layer String Kernel NN. Simply let $\psi^{(1)}_{i,j}=\phi_j^{(1)}(\mbf{w}_{i,j})$ completes the proof for the case of $l=1$.

Suppose the lemma holds for $l=1,\cdots,k$, we now prove the case of $l=k+1$. 
Similar to the proof of Theorem~\ref{theorem:onelayer}, the value of $\mbf{c}_j^{(k+1)}[t][i]$ equals to
\begin{align}
\mbf{c}^{(k+1)}_j[t][i]\ &=\ \sum_{1\leq i_1<\cdots<i_j\leq t} \ 
\inner{\mbf{w}^{(1)}_i}{\h^{(k)}[i_1]}\cdots\inner{\mbf{w}^{(j)}_i}{\h^{(k)}[i_j]}\ \cdot\ \lambda^{t-i_1-j+1} \label{eq:cti2}
\end{align}
where $\mbf{w}^{(j)}_i$ is the i-th row of the parameter matrix $\W^{(j)}$ of the $l$-th layer.
Note $\h^{(k)}[t][i]=\inner{\phi_\sigma^{(k)}(\x_{1:t})}{\psi^{(k)}_{\sigma,i,n}}$, we construct a matrix $\mbf{M}$ by stacking all $\{\psi^{(k)}_{\sigma,i,n}\}_i$ as the row vectors.\footnote{Note in practice the mappings $\phi^{(k)}_\sigma$ and $\psi^{(k)}_\sigma$ may have infinite dimension because the underlying mapping for the non-linear activation $\phi_\sigma()$ can have infinite dimension. The proof still apply since the dimensions are still countable and the vectors have finite norm (easy to show this by assuming the input $\x_i$ and parameter $\W$ are bounded.} 
We then can rewrite $\h^{(k)}[t]=\mbf{M}\,\phi^{(k)}_\sigma(\x_{1:t})$.
Plugging this into Eq~(\ref{eq:cti2}), we get
\begin{align*}
\mbf{c}^{(k+1)}_j[t][i]\ &=\ \sum_{1\leq i_1<\cdots<i_j\leq t} \ 
\inner{\mbf{w}^{(1)}_i}{\mbf{M}\,\phi^{(k)}_\sigma(\x_{1:i_1})}\cdots\inner{\mbf{w}^{(j)}_i}{\mbf{M}\,\phi^{(k)}_\sigma(\x_{1:i_j})}\ \cdot\ \lambda^{t-i_1-j+1} \\[6pt]
  &=\ \sum_{1\leq i_1<\cdots<i_j\leq t} \ 
\inner{\underbrace{\mbf{M}^\top\,\mbf{w}^{(1)}_i}_{\mbf{u}_{i,1}}}{\phi^{(k)}_\sigma(\x_{1:i_1})}\cdots\inner{\underbrace{\mbf{M}^\top\,\mbf{w}^{(j)}_i}_{\mbf{u}_{i,j}}}{\phi^{(k)}_\sigma(\x_{1:i_j})}\ \cdot\ \lambda^{t-i_1-j+1} \\[6pt]
  &=\ \sum_{1\leq i_1<\cdots<i_j\leq t} \ 
\inner{\mbf{u}_{i,1}}{\phi^{(k)}_\sigma(\x_{1:i_1})}\cdots\inner{\mbf{u}_{i,j}}{\phi^{(k)}_\sigma(\x_{1:i_j})}\ \cdot\ \lambda^{t-i_1-j+1} \\[6pt]
  &=\ \inner{\underbrace{\sum_{1\leq i_1<\cdots<i_j\leq t} \lambda^{t-i_1-j+1}\, \phi^{(k)}_\sigma(\x_{i_1})\otimes\cdots\otimes\phi^{(k)}_\sigma(\x_{i_j})}_{\phi^{(k+1)}(\x_{1:t})}}{\ \mbf{u}_{i,1}\otimes\cdots\otimes\mbf{u}_{i,j}} \\[6pt]
  &=\ \inner{\phi^{(k+1)}(\x_{1:t})}{\ \mbf{u}_{i,1}\otimes\cdots\otimes\mbf{u}_{i,j}}
\end{align*}
Define $\psi^{(k+1)}_{i,j}=\mbf{u}_{i,1}\otimes\cdots\otimes\mbf{u}_{i,j}$ completes the proof for the case of $l=k+1$. \qed

\clearpage
\section{Proof for Theorem \ref{theorem:graph-onelayer}}
Recall that random walk graph kernel is defined as:
\begin{eqnarray}
\kernel_n(G,G') &=& \lambda^n \sum_{\x \in P_n(G)} \sum_{\y \in P_n(G')} \prod_{i=1}^n \inner{\f_{\x_i}}{\f_{\y_i}} \nonumber \\
\phi_n(G) &=& \lambda^n \sum_{\x \in P_n(G)} \f_{\x_0}\otimes\f_{\x_1} \otimes\cdots\otimes\f_{\x_n} \nonumber
\end{eqnarray}
and single-layer graph NN computes its states as:
\begin{eqnarray}
\cc_n[v] &=& \lambda\sum_{u\in N(v)} c_{n-1}[u] \odot \W^{(n)} \f_v \nonumber \\
&=& \lambda^n \sum_{u_{n-1}\in N(v)}\sum_{u_{n-2}\in N(u_{n-1})}\cdots\sum_{u_0\in N(u_1)} \W^{(0)}\f_{u_0}\odot\W^{(1)}\f_{u_1}\odot\cdots\odot\W^{(n)}\f_{u_n} \nonumber \\
&=& \lambda^n \sum_{\mbf{u}=\mbf{u}_0\cdots\mbf{u}_n \in P_n(G,v)} \W^{(0)}\f_{u_0}\odot\W^{(1)}\f_{u_1}\odot\cdots\odot\W^{(n)}\f_{u_n} \nonumber
\end{eqnarray}
where we define $P_n(G,v)$ be the set of walks that end with node $v$. For the $i$-th coordinate of $\cc_n[v]$, we have
\begin{eqnarray}
\cc_n[v][i] &=& \lambda^n \sum_{\mbf{u}\in P_n(G,v)} \prod_i \inner{\w_i^{(i)}}{f_{\mbf{u}_i}} \nonumber \\
\sum_v\cc_n[v][i] &=& \lambda^n \sum_{\mbf{u}\in P_n(G)} \prod_i \inner{\w_i^{(i)}}{f_{\mbf{u}_i}} \nonumber \\
&=& \kernel_n(G,L_{n,k}) \nonumber
\end{eqnarray}
The last step relies on the fact that $P_n(L_{n,k})$ has only one element $\{\w_i^{(0)},\w_i^{(1)},\cdots,\w_i^{(n)}\}$.\qed

\clearpage
\section{Proof for Theorem~\ref{theorem:deep_graph_nn}}
For clarity, we re-state the kernel definition and theorem in the following:
\begin{equation}
\kernel^{(L,n)}(G,G') = \sum_v\sum_{v'} \kernel_{loc,\sigma}^{(L,n)}(v,v') \nonumber
\end{equation}
When $l=1$:
$$
\kernel_{loc}^{(l,j)}(v,v')=
\begin{cases}
\displaystyle
\ \inner{\f_v}{\f_{v'}} & \text{if } j=1 \\[0.5ex]
\ \inner{\f_v}{\f_{v'}} + \lambda \sum\limits_{u\in N(v)}\sum\limits_{u'\in N(v')} \kernel^{(l,j-1)}_{loc,\sigma}(u,u') & \text{if } j>1 \\
0 & \text{if } P_j(G,v)=\emptyset \text{ or } P_j(G',v')=\emptyset
\end{cases}
$$
When $l > 1$:
$$
\kernel_{loc}^{(l,j)}(v,v')=
\begin{cases}
\displaystyle
\ \kernel^{(l-1)}_{loc,\sigma}(v,v') & \text{if } j=1 \\[0.5ex]
\ \kernel^{(l-1)}_{loc,\sigma}(v,v') + \lambda \sum\limits_{u\in N(v)}\sum\limits_{u'\in N(v')} \kernel^{(l,j-1)}_{loc,\sigma}(u,u') & \text{if } j>1 \\
0 & \text{if } P_j(G,v)=\emptyset \text{ or } P_j(G',v')=\emptyset
\end{cases}
$$
where $P_j(G,v)$ is the set of walks of length $j$ starting from $v$. Note that we force $\kernel_{loc}^{(l,j)}=0$ when there is no valid walk starting from $v$ or $v'$, so that it only considers walks of length $j$. We use additive version just for illustration (multiplicative version follows the same argument).
\begin{theo}
For a deep random walk kernel NN with $L$ layers and activation function $\sigma(\cdot)$, assuming the final output state $\h_G=\sum_v\h^{(l)}[v]$, for $l=1,\cdots,L;j=1,\cdots,n$:
\begin{enumerate}
\item[(i)] $\mbf{c}^{(l)}_j[v][i]$ as a function of input $v$ and graph $G$ belongs to the RKHS of kernel $\kernel^{(l,j)}_{loc}(\cdot,\cdot)$ in that
$$
\mbf{c}^{(l)}_j[v][i] = \inner{\phi_G^{(l,j)}(v)}{\psi_{i,j}^{(l)}}
$$
where $\phi_G^{(l)}(v)$ is the mapping of node $v$ in graph $G$, and $\psi_{i,j}^{(l)}$ is a mapping constructed from model parameters.

\item[(ii)] $\h^{(l)}[v][i]$ belongs to the RKHS of kernel $\kernel^{(l,n)}_{loc,\sigma}(\cdot,\cdot)$ as a corollary: 
\begin{eqnarray}
\h^{(l)}[v][i] &=& \sigma\left(\sum_k u_{ik}\cc^{(l)}_n[v][k]\right) = \sigma\left(\sum_k u_{ik}\inner{\phi_G^{(l,n)}(v)}{\psi_{k,n}^{(l)}}\right) \nonumber \\
&=& \sigma\left(\inner{\phi_G^{(l,n)}(v)}{\sum_k u_{ik}\psi_{k,n}^{(l)}}\right) \nonumber \\
&=& \inner{\phi_\sigma(\phi_G^{(l,n)}(v))}{\psi_\sigma\left(\sum_k u_{ik}\psi_{k,n}^{(l)}\right)} \nonumber
\end{eqnarray}
We denote $\psi_\sigma\left(\sum_k u_{ik}\psi_{k,n}^{(l)}\right)$ as $\psi_{\sigma,i,n}^{(l)}$, and $\phi_\sigma(\phi_G^{(l,n)}(v))$ as $\phi_{G,\sigma}^{(l)}(v)$ for short.
\item[(iii)] $\h_G[i]$ belongs to the RKHS of kernel $\kernel^{(L,n)}(\cdot,\cdot)$.
\end{enumerate}
\end{theo}
\textbf{Proof of (i), (ii):} We prove by induction on $l$. When $l=1$, from kernel definition, the kernel mapping is recursively defined as:
$$
\phi_G^{(1,j)}(v)=
\begin{cases}
\displaystyle
\ \f_v & \text{if } j=1 \\[0.5ex]
\ \left[\f_v, \sqrt{\lambda}\sum\limits_{u\in N(v)}\phi_\sigma(\phi_G^{(1,j-1)}(u)) \right] &  \text{if } j>1 \\
\mbf{0} & \text{if } P_j(G,v)=\emptyset \text{ or } P_j(G',v')=\emptyset
\end{cases}
$$
We prove $\mbf{c}^{(1)}_j[v][i] = \inner{\phi_G^{(1,j)}(v)}{\psi_{i,j}^{(1)}}$ by induction on $j$. When $j=1$, $\mbf{c}^{(1)}_1[v][i]=\inner{\w_i^{(1,1)}}{\f_v}$. Our hypothesis holds by $\psi_{i,1}^{(1)}=\w_i^{(1,1)}$. Suppose induction hypothesis holds for $1,2,\cdots,j-1$. If $P_j(G,v)=\emptyset$, we could always set $\mbf{c}^{(1)}_j[v][i]=0$ in neural network computation. Otherwise:
\begin{eqnarray}
\mbf{c}^{(1)}_j[v][i] &=& \inner{\w_i^{(1,j)}}{\f_v}+\lambda\sum_{u\in N(v)} \sigma\left(\cc_{j-1}^{(1)}[u][i]\right) \nonumber \\
&=& \inner{\w_i^{(1,j)}}{\f_v}+\lambda\sum_{u\in N(v)} \sigma\left(\inner{\phi_G^{(1,j-1)}(u)}{\psi_{i,j-1}^{(1)}}\right) \nonumber \\
&=& \inner{\w_i^{(1,j)}}{\f_v}+\sqrt{\lambda}\sum_{u\in N(v)} \inner{\phi_\sigma(\phi_G^{(1,j-1)}(u))}{\sqrt{\lambda}\psi_\sigma(\psi_{i,j-1}^{(1)})} \nonumber \\
&=& \inner{\left[\f_v,\sqrt{\lambda}\sum_{u\in N(v)}\phi_\sigma(\phi_G^{(1,j-1)}(u))\right]}{\left[\w_i^{(1,j)},\sqrt{\lambda}\psi_\sigma(\psi_{i,j-1}^{(1)})\right]} \nonumber
\end{eqnarray}
Let $\psi_{i,j}^{(1)}=\left[\w_i^{(1,j)},\sqrt{\lambda}\psi_\sigma(\psi_{i,j-1}^{(1)})\right]$ concludes the base case $l=1$. Suppose induction hypothesis holds for $1,2,\cdots,l-1$. Note that when $l>1$:
$$
\phi_G^{(l,j)}(v)=
\begin{cases}
\displaystyle
\ \phi_\sigma(\phi_G^{(l-1)}(v)) & \text{if } j=1 \\[0.5ex]
\ \left[\phi_\sigma(\phi_G^{(l-1)}(v)), \sqrt{\lambda}\sum\limits_{u\in N(v)}\phi_\sigma(\phi_G^{(l,j-1)}(u)) \right] &  \text{if } j>1 \\
\mbf{0} & \text{if } P_j(G,v)=\emptyset \text{ or } P_j(G',v')=\emptyset
\end{cases}
$$Now we prove $\mbf{c}^{(l)}_j[v][i] = \inner{\phi_G^{(l,j)}(v)}{\psi_{i,j}^{(l)}}$ by induction on $j$. When $j=1$, 
\begin{eqnarray}
\mbf{c}^{(l)}_1[v][i]&=&\inner{\w_i^{(l,1)}}{\h^{(l-1)}[v]} \nonumber \\
&=&\sum_k \w_{ik}^{(l,1)}\inner{\phi_{G,\sigma}^{(l-1)}(v)}{\psi_{\sigma,k,n}^{(l-1)}} \nonumber \\
&=&\inner{\phi_G^{(l,1)}(v)}{\sum_k \w_{ik}^{(l,1)}\psi_{\sigma,k,n}^{(l-1)}} \nonumber
\end{eqnarray}
Let $\psi_{i,j}^{(l)}=\sum_k \w_{ik}^{(l,1)}\psi_{\sigma,k,n}^{(l-1)}$ completes the proof for $j=1$. Now suppose this holds for $1,2,\cdots,j-1$. Same as before, we only consider the case when $P_j(G,v)\neq \emptyset$:
\begin{eqnarray}
\mbf{c}^{(l)}_j[v][i]&=&\inner{\w_i^{(l,j)}}{\h^{(l-1)}[v]}+\lambda\sum_{u\in N(v)} \sigma\left(\cc_{j-1}^{(l)}[u][i]\right) \nonumber \\
&=&\sum_k \w_{ik}^{(l,j)}\inner{\phi_{G,\sigma}^{(l-1)}(v)}{\psi_{\sigma,k,n}^{(l-1)}} + \lambda\sum_{u\in N(v)}\sigma\left(\inner{\phi_G^{(l,j-1)}(u)}{\psi_{i,j-1}^{(l)}}\right) \nonumber \\
&=&\inner{\phi_{G,\sigma}^{(l-1)}(v)}{\sum_k \w_{ik}^{(l,j)}\psi_{\sigma,k,n}^{(l-1)}} + \lambda\sum_{u\in N(v)} \inner{\phi_{G,\sigma}^{(l,j-1)}(u)}{\psi_\sigma(\psi_{i,j-1}^{(l)})} \nonumber \\
&=&\inner{\phi_{G,\sigma}^{(l-1)}(v)}{\sum_k \w_{ik}^{(l,j)}\psi_{\sigma,k,n}^{(l-1)}} + \inner{\sqrt{\lambda}\sum_{u\in N(v)}\phi_{G,\sigma}^{(l,j-1)}(u)}{\sqrt{\lambda}\psi_\sigma(\psi_{i,j-1}^{(l)})} \nonumber \\
&=&\inner{\left[\phi_{G,\sigma}^{(l-1)}(v),\sqrt{\lambda}\sum_{u\in N(v)}\phi_{G,\sigma}^{(l,j-1)}(u) \right]}{\left[\sum_k \w_{ik}^{(l,j)}\psi_{\sigma,k,n}^{(l-1)},\sqrt{\lambda}\psi_\sigma(\psi_{i,j-1}^{(l)})\right]} \nonumber \\
&=&\inner{\phi_G^{(l,j)}(v)}{\psi_{i,j}^{(l)}} \nonumber
\end{eqnarray}
Let $\psi_{i,j}^{(l)}=\left[\sum_k \w_{ik}^{(l,j)}\psi_{\sigma,k,n}^{(l-1)},\sqrt{\lambda}\psi_\sigma(\psi_{i,j-1}^{(l)})\right]$ concludes the proof. \qed

\textbf{Proof for (iii):} We construct a directed chain $L_{n,k}=(V,E)$ from model parameters, with nodes $V=\{l_n,l_{n-1},\cdots,l_0\}$ and $E=\{(v_{i+1},v_i)\}$. $l_j$'s underlying mapping is $\psi_{\sigma,i,j}^{(L)}$. Now we have
\begin{equation}
\h_G[i] = \sum_v \h^{(L)}[v][i] = \sum_{v\in G} \inner{\phi_\sigma(\phi_G^{(L)}(v))}{\psi_{\sigma,i,n}^{(L)}} = \sum_{v\in G} \sum_{v\in L_{n,k}} \kernel_{loc}^{(L,n)}(v,l_n) = \kernel^{(L,n)}(G,L_{n,k}) \nonumber
\end{equation}
Note that we are utilizing the fact that $\kernel_{loc}^{(L,n)}(v,l_j)=0$ for all $j\neq n$ (because $P_n(L_{n,j},l_j)=\emptyset$). \qed